\documentclass[conference]{IEEEtran}
\usepackage[latin1]{inputenc}
\ifCLASSINFOpdf
  \usepackage[pdftex]{graphicx}
\else
  \usepackage[dvips]{graphicx}
\fi
\usepackage[cmex10]{amsmath}
\usepackage{amsfonts}
\usepackage{amssymb}
\usepackage{graphicx}
\usepackage{physics}
\usepackage{cite}
\usepackage{siunitx}
\usepackage{enumerate}
\usepackage{mhchem}
\usepackage{mathtools}
\usepackage{graphicx}
\usepackage{float}
\usepackage{xcolor}
\usepackage{mdframed}
\usepackage{csquotes}
\usepackage{booktabs}
\usepackage{multicol}
\DeclareSIUnit\year{a}
\DeclareSIUnit\clight{c}
\mdfdefinestyle{exercise}{
	backgroundcolor=black!10,roundcorner=8pt,hidealllines=true,nobreak
}
\usepackage{color}
\definecolor{tumblue}{rgb}{0, 0.4, 0.74}
\begin{document}
\title{Optimizing Passenger Comfort in Cost Functions for Trajectory Planning}
\author{\IEEEauthorblockN{Jean Elsner}
\IEEEauthorblockA{Technische Universit\"at M\"unchen\\
Email: jean.elsner@tum.de}}
\maketitle
\begin{abstract}
Current advances in the development of autonomous cars suggest that driverless cars may see wide-scale deployment in the near future. Research by both industry and academia is driven by potential benefits of this new technology, including reductions in fatalities and improvements in traffic and fuel efficiency as well as greater mobility for people who will or cannot drive cars themselves. A deciding factor for the adoption of self-driving cars besides safety will be the comfort of the passengers. This report looks at cost functions currently used in motion planning methods for autonomous on-road driving. Specifically, how the human perception of how comfortable a trajectory is, can be formulated within cost functions.
\end{abstract}
\section{Introduction}
General interest and research into autonomous cars has been growing rapidly in recent years, fueled by promises such as enhanced safety, more efficient traffic flow, greater mobility as well as decreased fuel consumption. While verification of the safety and reliability of a self-driving car is paramount, real world systems will eventually be judged by the amount of comfort they provide their passengers. Fully autonomous transportation systems will no longer allow the human passengers to control the vehicle. Therefore, it is up to the system to plan the motion of the vehicle in a way, that does not put too much stress on the passengers and does not decrease their level of comfort. Ignoring human factors and ergonomics in the design of self-driving systems may induce negative side-effects in passengers and ultimately lead to their rejection of the technology. 

This report discusses different criteria that may be used to quantify how comfortable a trajectory is as perceived by humans and reviews how these criteria are currently implemented inside the cost functions of trajectory planning algorithms.
\subsection{Motion planning for on-road driving}
Planning for autonomous on-road driving can generally be divided into three stages \cite{katrakazas2015}:
\begin{enumerate}
\item Finding a global path to the goal destination
\item Planning and selecting the best maneuver to perform
\item Finding the best trajectory to follow
\end{enumerate}
While maneuver planning is concerned with selecting the most suitable high-level action (e.g. overtaking or lane changes), the path planer provides a geometric path to the destination. Trajectory planning on the other hand (also referred to as motion planning) is the most low-level planner and is evaluated each iteration of the planning cycle (consisting of sensing, planing and execution).
\subsection{Cost Functions in Motion Planning}
During motion planning, a number of trajectories are generated each cycle. The cost function is used to evaluate each of these trajectories and select the optimal trajectory according to the cost function's objectives. Mathematically, the cost function can be formulated as 
\begin{align}
\label{cost-function}
J_C(x(t),u(t),t_0,t_f) =\;& \underbrace{\Phi_C(x(t_0),t_0,x(t_f),t_f)}_{\text{terminal cost}} +\\ &\underbrace{\int_{t_0}^{t_f} L_C(x(t),u(t),t)\dd t.}_{\text{running \nonumber cost}}
\end{align}
With $x(t)$ the vehicle's state vector, $u(t)$ the input vector and $t_0$ and $t_f$ the initial and final time of the trajectory. The cost function can further be generalized to consist of two summands, the first of which depends only on the initial and final state and time. The second summand depends on state and input over the entire trajectory. To find the best trajectory the cost function will be minimized
\begin{align}
\label{ustar}
u^*(\cdot) = \underset{u(\cdot)}{\operatorname{arg\,min}}\; J_C(x(t),u(t),t_0,t_f)
\end{align}
subject to the constraints
\begin{align}
&\dot{x}(t) = f_M(x(t),u(t)),\qquad O(x(t))\in \mathcal{W}_{\text{S,free}},\\
&g_\text{S}(x(t),u(t),t)\leq 0,\qquad x(t_0),\qquad x(t_f)\in \mathcal{G}_\mathrm{S}.\nonumber
\end{align}
Where the first constraint defines the vehicle's dynamics and therefore enforces feasible solutions. The second one concerns the vehicle's occupancy $O(x(t))$, effectively avoiding collisions. Additional hard constraints, like speed limits or other traffic rules, can be encoded in $g_\text{S}(x(t),u(t),t)$. Finally, the trajectory must begin at the vehicle's current position given by $x(t_0)$ and end in an accepted goal region $\mathcal{G}_\text{S}$.
\subsection{CommonRoad}
The notation for cost functions in this report follows the convention introduced in the \underline{com}posable benchmarks for \underline{m}otion planning \underline{on} \underline{road}s (CommonRoad) framework \cite{althoff2017}. CommonRoad provides ready made vehicle dynamics, traffic scenarios and cost functions. A combination of these fully defines a motion planning benchmark that can be identified by a unique ID and therefore becomes easy to reproduce.

In order to construct arbitrary cost functions, the framework provides a set of both running and terminal basis cost functions, which have been listed in Table \ref{partials}. A cost function of the form seen in equation (\ref{cost-function}) can then be written as a weighted superposition of these partial cost functions:
\begin{align}
J_C(x(t),u(t),t_0,t_f) = \sum_{i\in\mathcal{I}}w_i J_i(x(t),u(t),t_0,t_f).
\end{align}
A shorthand is introduced to describe cost functions by the indices and weights of the partial cost functions used to construct them:
\begin{align}
[(X_1\mid w_1),...,(X_N\mid w_N)] = w_1 J_{X_1} + ... + w_N J_{X_N}.
\end{align}

\begin{table}[]
\centering
\caption{Partial cost functions in CommonRoad}
\label{partials}
\begin{tabular}{@{}ll@{}}
\toprule
\multicolumn{2}{l}{Running costs}                                                           \\ \midrule
Acceleration              & $J_A=\int_{t_0}^{t_f}a^2\dd t$                                  \\
Jerk                      & $J_J=\int_{t_0}^{t_f}\dot{a}^2\dd$                              \\
Steering angle            & $J_{SA}=\int_{t_0}^{t_f}\delta^2\dd t$                          \\
Steering rate             & $J_{SR}=\int_{t_0}^{t_f}a_\delta^2\dd t$                        \\
Energy                    & $J_E=\int_{t_0}^{t_f}P(x,u)^2\dd t$                             \\
Yaw rate                  & $J_Y=\int_{t_0}^{t_f}\dot{\Psi}^2\dd t$                         \\
Lane center offset        & $J_{LC}=\int_{t_0}^{t_f}d^2(t)\dd t$                            \\
Velocity offset           & $J_V=\int_{t_0}^{t_f}(v_{\text{des}}(x(t))-v(t))^2\dd t$           \\
Orientation offset        & $J_O=\int_{t_0}^{t_f}(\theta_{\text{des}}(x(t))-\theta(t))^2\dd t$ \\
Distance to obstacles     & $J_D=\int_{t_0}^{t_f}\operatorname{max}(\xi_1,...,\xi_0)\dd t$  \\
Path length               & $J_L=\int_{t_0}^{t_f}v\dd t$                                    \\ \midrule
\multicolumn{2}{l}{Terminal costs}                                                          \\ \midrule
Time                      & $J_T=t_f$                                                       \\
Terminal offset           & $J_{TO}=d^2(t_f)$                                        \\
Terminal distance to goal & $J_{TG}=d^2_{\text{goal}}(t_f)$                                
\end{tabular}
\end{table}
\section{Human Comfort Criteria}
\begin{figure}[htp!]
\begin{center}
\includegraphics[width=.5\textwidth]{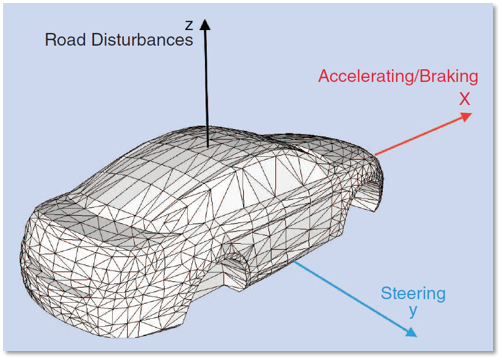}
\end{center}
\caption{Illustration of the resulting forces acting on a passenger within a vehicle and their root cause \cite{elbanhawi2015}. Note that only the horizontal forces are subject to be influenced by driver behavior.}
\label{forces-pic}
\end{figure}

\begin{figure}[htp!]
\begin{center}
\includegraphics[width=.5\textwidth]{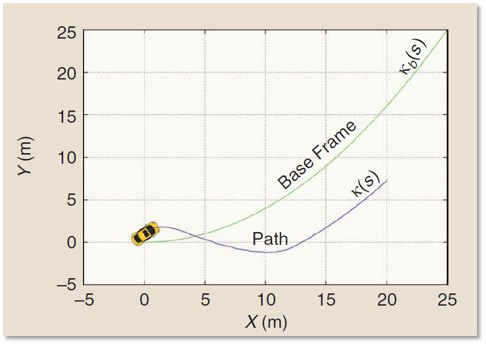}
\end{center}
\caption{Smooth candidate trajectory next to base path \cite{arnay2016}. Path planners will often generate a bunch of candidate trajectories that are parallel to the base path in a space where the base path is a straight line and rank them in euclidean space using cost functions.}
\label{curvilinear-pic}
\end{figure}

Most commonly, passenger comfort is taken into account by minimizing the vehicle's motion in regard to acceleration, jerk or both \cite{elbanhawi2015,zanasi2002}. While vertical forces act due to road disturbances as illustrated in figure \ref{forces-pic}, horizontal forces are the direct result of acceleration and steering and can therefore be considered by the system's motion planning algorithm. Within the CommonRoad framework, this approach may be quantified in the context of a cost function as
\begin{align}
J_\text{comf}(x(t),u(t),t_0,t_f)=\;&w_1 J_A+w_2 J_J=\\&w_1\int_{t_0}^{t_f}a^2\dd t + w_2\int_{t_0}^{t_f}\dot{a}^2\dd t.\nonumber
\end{align}
Or, using the shorthand, as
\begin{align}
[(A\mid w_1),(J\mid w_2)].
\end{align}\noindent
\noindent
Another method occasionally used is the selection of smoother paths. It is assumed that smoother paths are more natural and more closely resemble the behavior of human drivers \cite{elbanhawi2015}. Often, this is combined with a path planer that generates continuous curvilinear trajectory candidates, as illustrated in figure \ref{curvilinear-pic}.
\noindent
One way to penalize non-smooth trajectories in the cost function, is to minimize curvature. While there is no partial cost function for curvature in the CommonRoad framework, a respective cost function can be derived using the geometric definition:
\begin{align}
J_{\kappa}=\operatorname{max}\qty{\frac{x'_i\cdot y''_i-x''_i\cdot y'_i}{(x'_i+y'_i)^{3/2}}}
\end{align}
Here, the maximum curvature of a trajectory is considered by calculating the curvature at each point $(x_i\mid y_i)$ on the trajectory. Note however, that a more convenient method may be used to compute the curvature, if the candidate is a curvilinear function such as, for example, a clothoid.\newpage\noindent
The criteria for passenger comfort discussed so far are of a physical nature. Due to the loss of control however, human passengers in a driverless car may experience discomfort due to psychological effects, such as a perceived lack of safety \cite{beggiato}. For a motion planning system with a given response ratio
\begin{align}
\frac{vT}{d},
\end{align}
where $v$ is the current velocity and T is the response time (given by the period of the planning cycle), a minimum distance $d$ to obstacles must be kept, such that the system can respond quickly enough to changes in the environment so as to drive safely. Because a computer's responsiveness is orders of magnitude higher than a human's, a motion planning system might choose trajectories, that are perfectly safe for the system to execute, but would be impossible to navigate safely by a human driver. Because of this perceived lack of safety and the anxiety it may induce in human passengers, the proximity to obstacles may be artificially decreased by the motion planner. This can be quantified using the partial cost function
\begin{align}
J_D(x(t),u(t),t_0,t_f) = w_1 \int_{t_0}^{t_f}\operatorname{max}(\xi_1,...,\xi_o)\dd t=[(D\mid w_1)].
\end{align}
\section{State of the art}
This section discusses the state of the art of passenger comfort criteria in cost functions for motion planning systems by example of several publications. An attempt is made to formulate cost functions according to the conventions of the CommonRoad framework, were applicable.

\begin{figure}[htp!]
\begin{center}
\includegraphics[width=.24\textwidth]{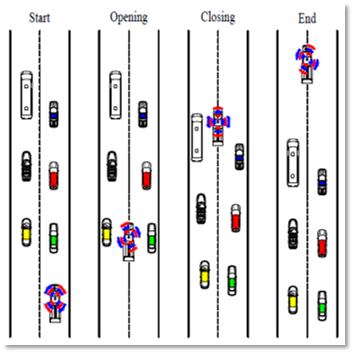}
\includegraphics[width=.24\textwidth]{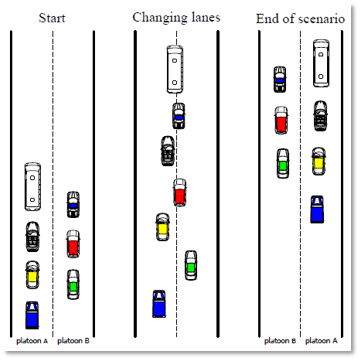}
\end{center}
\caption{Two examples of cooperative maneuvers presented in Mohseni 2017. The left illustration shows the for phases involved in creating a moving emergency passage. On the right a synchronized lane change is shown.}
\label{zipper}
\end{figure}

Mohseni 2017 presents fuel and comfort efficient control for autonomous vehicles \cite{mohseni2017}. The optimization of passenger comfort and fuel consumption is explicitly taken into account by the cost function. Non-comfort is defined as high jerk and acceleration. The average fuel power consumed by the engine is given as
\begin{align}
P = \frac{Fv}{\eta}
\end{align}
where $\eta$ is the engine efficiency, $F$ and $v$ are force and velocity respectively. This power is used to approximate the fuel flow
\begin{align}
V = \frac{P}{H\rho}\approx Fv.
\end{align}
With $\rho$ and $H$ the fuel density and fuel lower heating value respectively. With this, it is reasoned, that optimal fuel efficiency can be achieved by minimizing the engine power. The overall cost function can be written as
\begin{align}
J_{FM1} = [(A\mid w_1),(J\mid w_2),(E\mid w_2)].
\end{align}
Here, the CommonRoad convention to use the initials of the first author and a running number has been used to index the cost function. Note however, that no weights are given in Mohseni 2017. It is hinted, that they may even use different weights for the $x$ and $y$ components in global coordinates within the cost function. While no further evaluation of passenger comfort is made, it is worth mentioning, that Mohseni 2017 apply their cost function to several vehicles so as to plan cooperative maneuvers, such as synchronized lane changes or the zipper-like opening of a passage for an emergency vehicle (illustrated in figure \ref{zipper}).

\begin{figure}[htp!]
\begin{center}
\includegraphics[width=.5\textwidth]{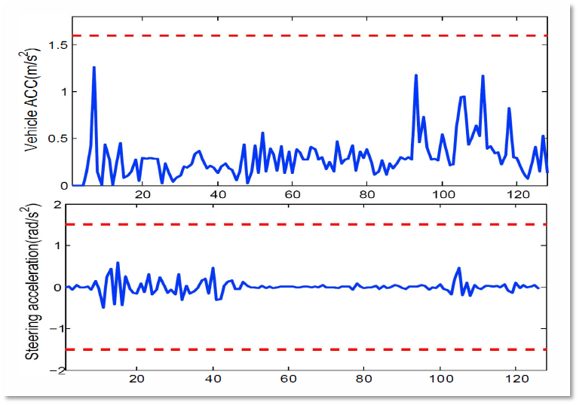}
\end{center}
\caption{Vehicle and steering acceleration of a vehicle controlled by the method proposed in Du 20016 \cite{du2016}. The red dashed lines represent safety and comfort boundaries that may not be crossed.}
\label{isojerk}
\end{figure}

\noindent
In Du 2016 a nonlinear model predictive control (NMPC) scheme is used to control the velocity and steering of a vehicle \cite{du2016}. Genetic algorithms are utilized to optimize a cost function, that is designed specifically to take the safety and comfort of human passengers into account. The cost function includes costs for tracking errors, control smoothness as well as high acceleration. While the cost function in Du 2016 is given in L1-norm and sums over discrete points on the trajectory up to the control or prediction horizon, here the cost function is written using the established CommonRoad conventions. 
\begin{align}
J_{XD1} =\;& [(LC\mid w_1),(O\mid w_2), (SR\mid w_3),(A\mid w_5(t))] \;+ \\ &\int_{t_0}^{t_f}w_4 a_{tan}^2+w_6(t)\operatorname{sgn}({a_{tan}})a_{tan}^2\dd t\;+\nonumber\\&\int_{t_0}^{t_f}w_7(t)(v_{max}-v)^2\dd t\nonumber 
\end{align}
With the last 3 weights defined as
\begin{align}
w_5(t) =\;&\begin{cases}
0,\qquad \text{if } a \leq a_{max}\\
w_{5,0} \qq{otherwise}
\end{cases}
\\
w_6(t) =\;&\begin{cases}
w_{6,0},\qq{if condition1}\\
0, \qq{otherwise}
\end{cases}
\\
w_7(t) =\;&\begin{cases}
0,\qq{if condition1}\\
w_{7,0}, \qq{otherwise}
\end{cases}\\\label{condition}
\text{condition1} =\;& (a_{tan} > 0) \land ((a > a_{max})\; \lor\\
&(d(t)> \SI{0.3}{\meter})\lor((\theta_{des}(x(t))-\theta(t))>\num{0.09}))\nonumber
\end{align}
In dangerous situations as identified in equation (\ref{condition}), forward acceleration is penalized, while under normal conditions, the vehicle is supposed to drive at a prescribed cruise speed of $v_{max}$. Furthermore, minimizing the steering rate is thought to prevent sudden and large changes that may induce shaking, while horizontal acceleration is penalized so as to follow the ISO 2631-1 standard. Du 2016 argues that passengers will not feel any shaking or jerking effects, as steering acceleration and vehicle acceleration during their tests were confined to comfort regions as illustrated in figure \ref{isojerk}.
\begin{figure*}[htp!]
\begin{center}
\includegraphics[width=1\textwidth]{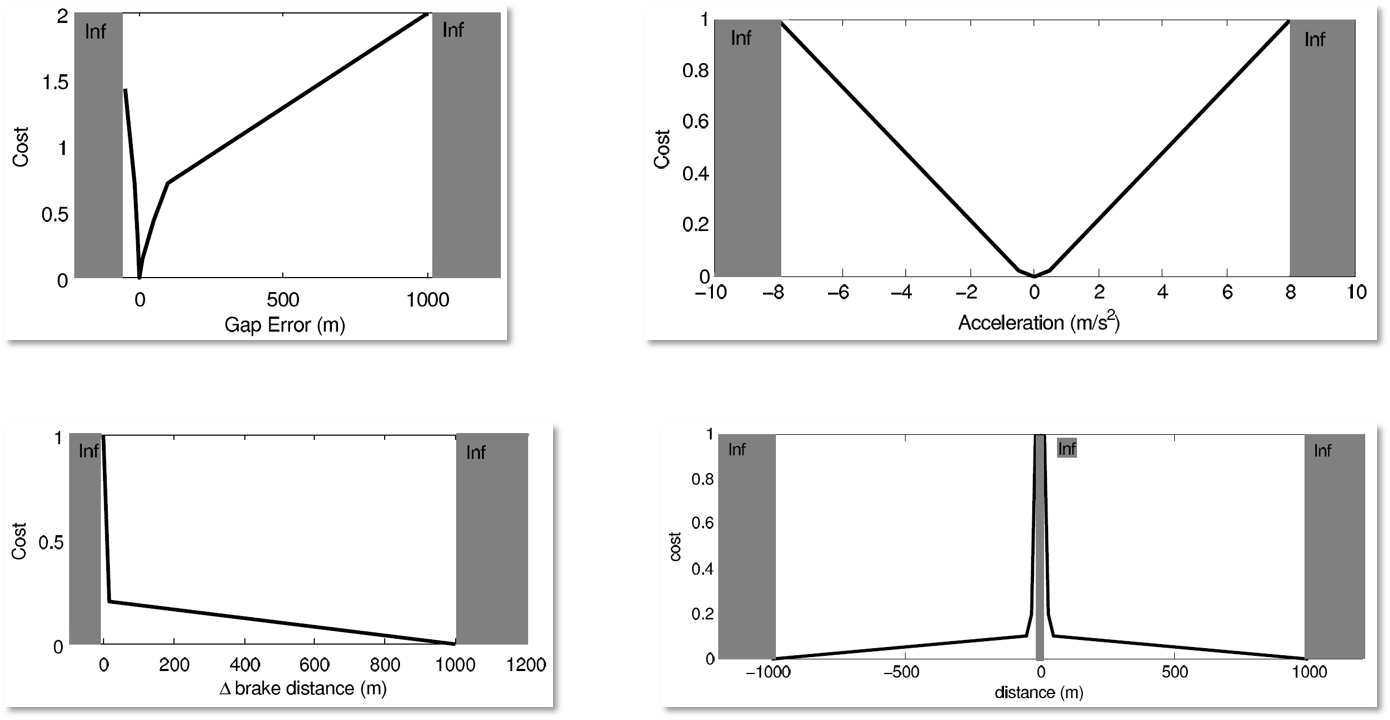}
\end{center}
\caption{Cost function library used in Wei 2016 \cite{du2016}. Each cost function consists of a sequence of connected vertices with the cost outside defined as infinite. From left to right and top down the cost functions are the distance keeper progress cost, comfort cost, braking distance cost and clear distance cost respectively.}
\label{costlib}
\end{figure*}\noindent
Wei 2010 uses a prediction and cost function based algorithm to implement autonomous freeway driving \cite{wei2010}. A library of cost functions, consisting of progress, comfort and safety costs, is used to evaluate the strategies generated by the three modules \emph{distance keeper}, \emph{lane selector} and \emph{merge planner}. While the latter two modules are concerned with maneuver planning, the former is a sort of lower level cruise control, that follows the leading vehicle. If no leading vehicle is detected, a virtual vehicle driving at the speed limit is inserted $\SI{150}{\meter}$ ahead. The cost functions used in Wei 2010 are designed to be human-understandable and informative. They consist of piecewise linear functions, while unacceptable inputs are mapped to infinite costs, as pictured in figure \ref{costlib}. For the \emph{distance keeper}, the cost function consists of a progress, safety and comfort cost. The progress cost is defined as the difference between the current distance to the leading vehicle and the desired distance to it (a function of velocity), called the gap error in figure \ref{costlib}. The safety cost penalizes proximity to other vehicles as well as the braking distance to the leading vehicles. These two costs are evaluated according to the $\Delta \text{ brake distance }$ and distance costs of figure \ref{costlib}. Finally, the comfort part of the cost functions in Wei 2010 corresponds to the acceleration cost in figure \ref{costlib}. Introducing a partial cost function for the distance to the leading vehicle as
\begin{align}
J_{LV} = \int_{t_0}^{t_f}(d_{l,des}-d_{l})^2\dd t.
\end{align}
With the desired distance to the leading vehicle $d_{l,des}=d_{l,min}+k_{gain}v$ and the current distance to the leading vehicle $d_l$. As well as a partial cost function for the brake distance to the leading vehicle
\begin{align}
J_{BD} = \int_{t_0}^{t_f}(d_l+\frac{1}{2}v_l^2a_{maxdec}^{-1}-vT-\frac{1}{2}v^2a_{maxdev}^{-1})^2\dd t.
\end{align}
Where $v_l$ is the velocity of the leading vehicle, $a_{maxdec}$ the maximum deceleration and $T$ is the motion planner's response time. Allows us to introduce the cost function
\begin{align}
J_{JW1} = [(LV\mid 50),(A\mid 10),(BD\mid 30),(D\mid 20)].
\end{align}
With $k_{gain}$ given in \cite{wei2010} as \SI{1.14}{\second}, $d_{l,min}=\SI{5}{\meter}$ and $T=\SI{.6}{\second}$. Note, that this formulation uses an L2 norm to calculate the running costs and ignores the piecewise functional forms given in Wei 2010. The regions with infinite cost however, can be trivially checked within the hard constraints $g_s(x(t),u(t),t)\leq 0$.
Arnay 2016 introduces a local planner, that produces a set of candidate trajectories, which are scored using a linear combination of weighted cost functions \cite{arnay2016}. An attempt is made, to discern the influence of the different weights on the prototype's behavior and find the optimal set of weights. Candidate local paths are generated by transforming the euclidean coordinate system into Fren\'et space, with the global path as the base frame. Trajectories are computed in curvilinear space and then transformed to the original euclidean space, where the cost is computed.
\begin{figure}[htp!]
\begin{center}
\includegraphics[width=.5\textwidth]{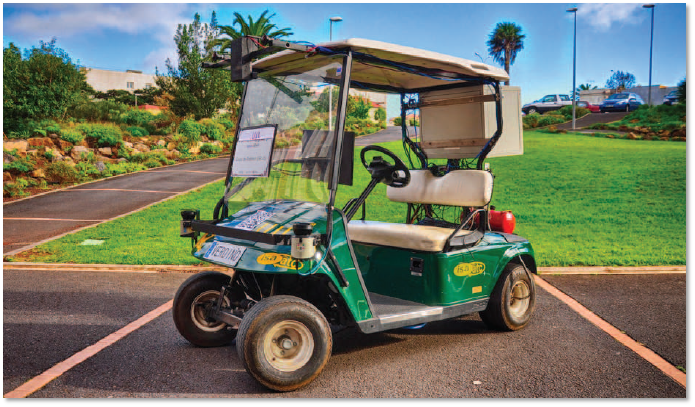}
\end{center}
\caption{The Verdino platform on which the method presented in Arnay 2016 is tested \cite{arnay2016}. The vehicle is a golf car that has been modified to be controlled by an on-board computer. It can reach a speed of around $\SI{20}{\kilo\meter\per\hour}$ and is equipped with several sensors for mapping and localization.}
\label{verdino}
\end{figure}\noindent
Before the cost function can be formulated in CommonRoad notation, an additional partial running cost has to be introduced
\begin{align}
J_C = \omega_c \frac{1}{s_2-s_1}\int_{s_1}^{s_2}l_i\dd s.
\end{align}
$J_C$, referred to as the consistency cost, penalizes changing the trajectory candidate between planning cycles. Where $l_i(s)$ is the lateral distance between the current and the previous chosen trajectory at the same longitudinal position parameterized by $s$. $s_1$ and $s_2$ are the first and last positions over $s$ where the trajectories share points. Given this and the previously introduced curvature cost, the cost function can be written as
\begin{align}
J_{RAi} =& [(D\mid \omega_o),(L\mid \omega_l),(LC\mid \omega_d),(\kappa\mid \omega_k),(C\mid \omega_c)].
\end{align}
The curvature cost is chosen to select smooth paths and consider passenger comfort. While horizontal forces and jerk are not considered explicitly, it's worth mentioning, that the motion planning system is run on a modified golf car with only a 36 $V_{CC}$ electrical motor that achieves a maximum speed between 19 and 25 \si{\kilo\meter\per\hour} (cf. figure \ref{verdino}).
\begin{figure*}[htp!]
\begin{center}
\includegraphics[width=\textwidth]{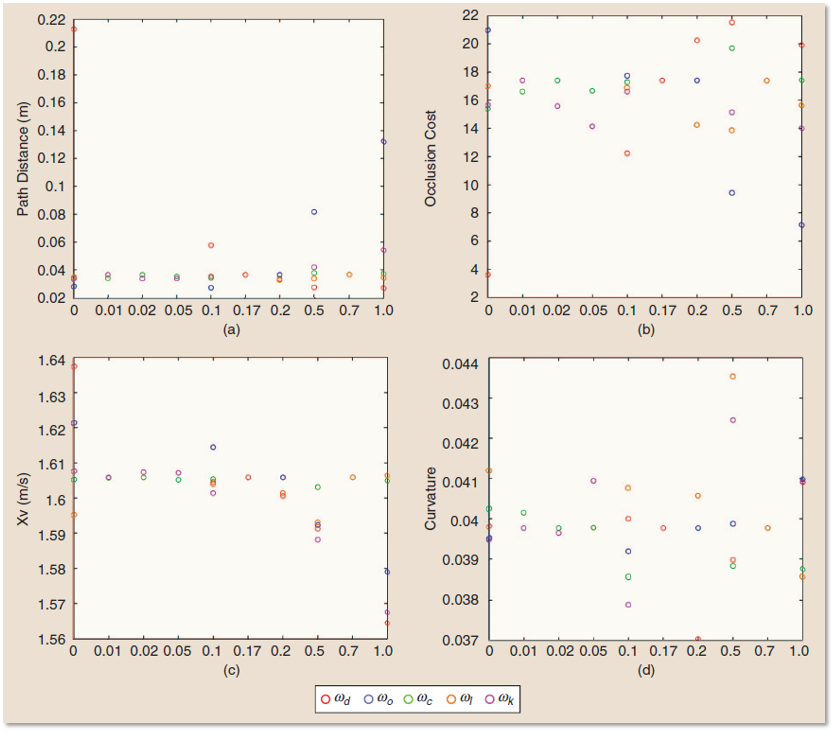}
\end{center}
\caption{Influence of single weights on choosing trajectories in the path planner presented in Arnay 2016 \cite{arnay2016}. The measurements are taken by clamping all the other weights while varying one weight in on the interval $[0-1]$. The x-axis defines the value assigned to the varied weight while the y-axis shows the influence on the 4 measured metrics path distance, occlusion, speed and curvature respectively.}
\label{weights}
\end{figure*}
\begin{figure}[htp!]
\begin{center}
\includegraphics[width=.35\textwidth]{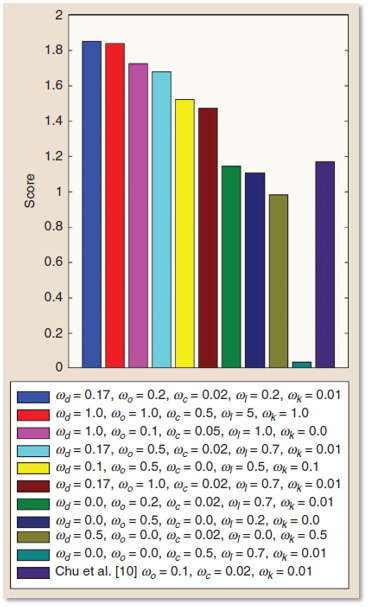}
\end{center}
\caption{Ranking of different tuples of weight values scored according to distance to obstacles and proximity to lane center in Arnay 2016 \cite{arnay2016}.}
\label{ranks1}
\end{figure}
\begin{figure}[htp!]
\begin{center}
\includegraphics[width=.35\textwidth]{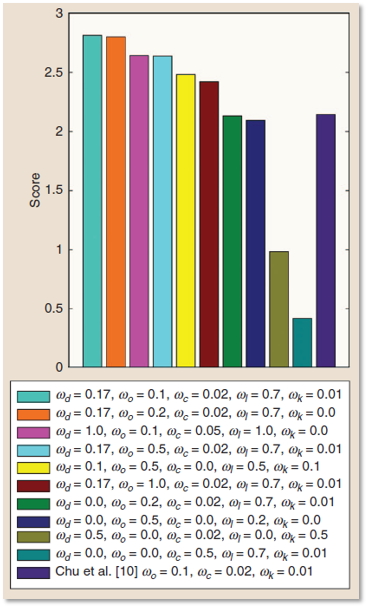}
\end{center}
\caption{Ranking of different tuples of weight values according to distance to obstacles, proximity to lane center, speed and the inverse of curvature in Arnay 2016 \cite{arnay2016}.}
\label{ranks2}
\end{figure}\noindent
In order to evaluate the relative importance of each weight, a simulation was set up, where the start and goal position as well as the positions of the obstacles were kept the same. Starting from a base configuration of $\omega_d=0.17$, $\omega_o=0.2$, $\omega_c=0.02$, $\omega_l=0.7$, $\omega_k=0.01$, which was found empirically, each of the weights was varied between 0 and 1, while the other weights kept their default value. The resulting trajectories of the simulation are then evaluated in regard to their lane center offset (a), distance to obstacles (b), speed(c) and curvature(d). The results can be seen in Fig. \ref{weights}. As would be expected, the measured variables of the trajectories correlate more strongly with their corresponding cost weight. For example, variation of the lane center offset weight $w_d$ is inversely proportional to the lance center cost of the resulting trajectory in figure \ref{weights} (a), whereas the same weight is proportional to the distance to obstacles cost in \ref{weights} (b). The latter might be explained by the fact, that forcing the system to keep close to the global path as much as possible, might result in proximity to obstacles not considered during geometric path generation. The opposite can be observed for the distance to obstacles weight $\omega_o$. The more influence the weight gets, the costlier the trajectories become in regard to lane center offset. This seems also sensible: if the vehicle's priority is to give obstacles a wide berth at all costs, it may strongly deviate from its path. The weights' influence on the speed cost are more specific to the motion planner used in Arnay 2016, as speed commands are e.g. computed based on the inverse of the length of the path and proximity to obstacles. No discernible pattern can be determined for the curvature cost, all included weights seem to influence it concurrently, with no one weight strictly dominant.

Finally, different sets of weights are ranked as seen in figure \ref{ranks1} and figure \ref{ranks2}. The ranking in figure \ref{ranks1} considers proximity to obstacles and lane center offset, while the one seen in Fig. \ref{ranks2} also takes speed and path curvature into account. For reference, the weights used in Chu 2012 are added to the ranking \cite{chu2012}. The comparison however does not seem fair, as Chu 2012 does not include weights for lane center offset as well as path length, both of which are considered in the ranking, albeit path length only indirectly via speed. This is especially true, as the weights are otherwise nearly the same. The highest ranking weights for the two respective rankings in figure \ref{ranks1} and figure \ref{ranks2} for Arnay 2016 can be summarized using CommonRoad notation as
\begin{align}
J_{RA1} =\;& [(D\mid 0.2),(L\mid 0.2),(LC\mid 0.17)]+\\
&[(\kappa\mid 0.01),(C\mid 0.02)]\nonumber\\
J_{RA2} =\;& [(D\mid 0.1),(L\mid 0.7),(LC\mid 0.17)]\\
&[(\kappa\mid 0.01),(C\mid 0.02)].\nonumber
\end{align}
The cost function in \cite{chu2012} accordingly can be written as
\begin{align}
J_{KC1} =\;& [(D\mid 0.1),(\kappa\mid 0.01),(C\mid 0.02)].
\end{align}
Note that even though curvature and therefore smoothness of the trajectories was specifically mentioned as a means to consider passenger comfort in Arnay 2016, most of the final results assign only a small weight or none at all to curvature.
\section{Conclusion}
This report presented a review of cost functions employed in current motion planning systems in regard to passenger comfort. It was established, that most approaches are concerned with directly limiting resulting forces and jerk on the passenger. While there are sometimes references to norms concerning exposure of the human body to vibration and shock, such as ISO-2631-1 \cite{katrakazas2015}, \cite{du2016}, \cite{elbanhawi2015}, \cite{gonzales2016}, there are few reports validating their assumptions in studies including live test subjects. Side effects such as motion sickness or anxiety may have very subjective root causes, that are difficult to quantify \cite{elbanhawi2015}, \cite{beggiato}. Because of this, approaches outside of the cost function may proof fruitful. In Whitsitt 2012 an upper bound for the forward velocity in relation to the steering angle was derived from data captured from human drivers \cite{whitsitt2012}. As an extension, even more complex control relations could be learned using methods from machine learning. Another research direction, seen in Gonz\'ales 2016 directly generates smooth speed profiles using quintic B\'ezier curves. In the end, more field tests with live subjects under real conditions are required to identify the needs of passengers more precisely. Eventually it may turn out, that a robust and safe motion planner, that obeys all traffic rules, is inherently comfortable for the majority of people.

\end{document}